%% file: article.tex
\documentclass{wielgosz-info-article}

\pdfoutput=1

\papertype{Concept paper}
\title{Falls Prediction in eldery people using Gated Recurrent Units}
\author{Marcin Radzio and Maciej Wielgosz and Matej Mertik}

\begin{document}
\begin{titlepage}
    \begin{authblk}{Marcin Radzio}{}
        AGH University of Science and Technology, Krak\'ow, Poland
    \end{authblk}

    \begin{authblk}{Maciej Wielgosz}{0000-0002-4401-2957}
        Academic Computer Centre CYFRONET AGH, Krak\'ow, Poland
    \end{authblk}
    
    \begin{authblk}{Matej Mertik}{0000-0002-4557-330X}
        Alma Mater Europaea ECM, Slovenia
    \end{authblk}

    \maketitle

    \begin{abstract}
        \input{abstract}
    \end{abstract}
    
    
    \begin{correspondence}
        Maciej Wielgosz <wielgosz@agh.edu.pl>
    \end{correspondence}
\end{titlepage}


\nocite{radzio2019falls}
\input{contents}

\input{ack}

\bibliography{bibliography,wielgosz-published}
\end{document}

%% file: abstract.tex
Falls prevention, especially in older people, becomes an increasingly important topic in the times of aging societies. In this work, we present Gated Recurrent Unit-based neural networks models designed for predicting falls (syncope). The cardiovascular systems signals used in the study come from Gravitational Physiology, Aging and Medicine Research Unit, Institute of Physiology, Medical University of Graz. We used two of the collected signals, heart rate, and mean blood pressure. By using bidirectional GRU model, it was possible to predict the syncope occurrence approximately ten minutes before the manual marker.

%% file: keywords.tex
GRU; RNN; syncope prediction

%% file: contents.tex
\input{contents-00-introduction}
\input{contents-20-experiments}
\input{contents-40-conclusions-future}

%% file: contents-00-introduction.tex
\section{Introduction}
\label{section:intro}

Patients residing in hospitals, especially elders, tend to weaken over time and are prone to never fully recovering \cite{martinezvelilla2015functional}.
The adequate exercise regime during hospitalization can stop or even reverse that effect \cite{martinezvelilla2019effect}.
Sometimes, however, during the remobilization, the patients happen to faint.
After undergoing the syncope, patients experience additional stress, which then results in a lack of self-confidence and undermines trust in the rehabilitation process.
Often it could be avoided by predicting whether the person is likely to faint.
Such an individual could be commissioned to further rehabilitation while not being exposed to stress associated with the collapse.

When it comes to falls prediction, there were several systems proposed \cite{rajagopalan2017fall}. Among others, there were attempts to asses the syncope risk by applying the ML (and isolation forest in particular) to the outcome of cognitive and motor tests \cite{mateen2016machine}, or use the accelerometer data gathered by the wearable sensor \cite{naitaicha2018deep}.

The purpose of this work is to develop a model based on recurrent neural networks that would be capable of forecasting occurrences of syncope by using real-life cardiological time series.
The collapse ideally should be predicted ahead of time, allowing the technician to interrupt the examination.
The models should also be highly sensitive -- patients who are certainly going to faint must be classified correctly, even at the cost of the overall accuracy.

The section~\ref{section:data} of the paper provides information about the dataset and preprocessing. Quality measures are described in section~\ref{section:quality_measures}, and the experimental setup and results are presented in section~\ref{section:experiments}. Finally, the conclusions of our research are presented in section \ref{section:conclusions}.

%% file: contents-20-experiments.tex
\section{Dataset}
\label{section:data}

The provided data consisted of nearly 700 files. Each file was labeled either as ,,syncope'' (indicating fainting during examination) or ,,no findings'', which for simplicity we will call ,,nosyncope'' from now on. For each patient, several measurements were performed, which upon further investigation turned out to be strongly correlated. After consultations, we used only two signals (with sampling rate \SI{1.25}{\hertz}) in further tests:
\begin{itemize}
    \item \texttt{mBP} -- mean blood pressure, and
    \item \texttt{HR} -- heart rate.
\end{itemize}

Since the selected data was incomplete, unbalanced, and fuzzy, it needed further cleaning. In a few cases, data happened to be wrongly labeled or existed in both classes.
For some time, no one noticed this issue.
As a result, we trained, evaluated, and tested the models on erroneous data, which caused their low predictive capabilities and worse overall accuracy.
The problem was eventually solved by fixing the labels and removing the redundant files.

We trimmed the first 500 and the last 50 samples of every time series, as their collecting happened during the start, fine-tuning or stopping the measuring devices.
After the trimming, we removed the series shorter than 500 samples, since they were deemed unusable for model training.

\begin{figure}
    \centering
    \includegraphics[width=\textwidth]{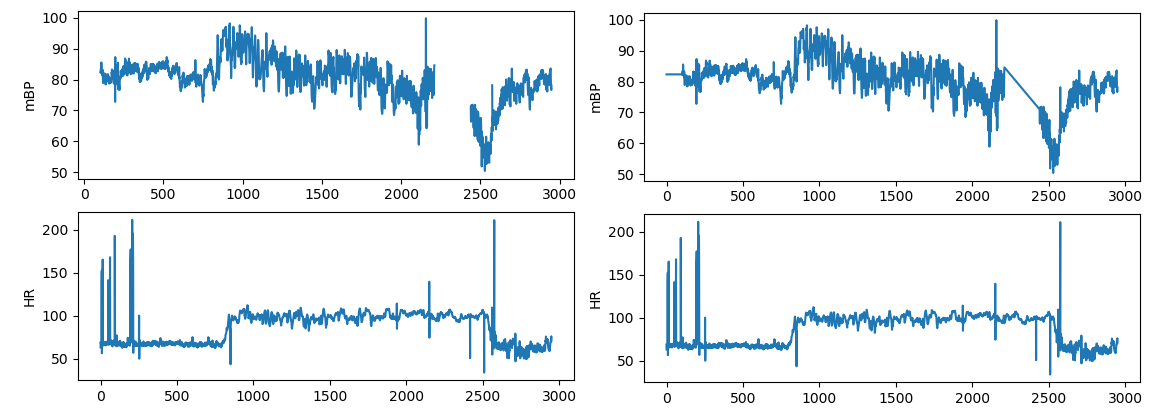}
    \caption{Example signals before and after interpolation.}
    \label{fig:experiments:dataset:data}
\end{figure}

Often, there were gaps in the signals, and the \texttt{mBP} and \texttt{HR} started and ended independently from each other. Depending on the gap location, we followed one of the two scenarios:
\begin{itemize}
    \item when the missing data was at the start or the end of the series, the first or last available data value, respectively, was used to fill in the gap, or
    \item when the gap was in the middle of the signal, the linear interpolation was used.
\end{itemize}
\noindent
Please see Fig.~\ref{fig:experiments:dataset:data} for the example data series before and after interpolation.

We devised an iterative procedure composed of several steps to remove the outliers from the signal:
\begin{enumerate}
    \item perform input signal studentization,
    \item apply the median filter (window size \num{31}),
    \item find difference between the studentized signal and the median filter output,
    \item identify outliers by comparing the difference to the threshold value,
    \item remove outliers from the input signal and interpolate the gaps.
\end{enumerate}
\noindent
The threshold value decreased with each iteration, and \numrange{2}{5} iterations were sufficient to clean even very noisy signals.

As a final preprocessing step, the data was normalized using the minmax normalization and rescaled to $[-1,1]$ range.

The ratio of ,,nosyncope'' to ,,syncope'' series in the dataset was almost 6:1.
In order to use the classification approach, we applied the data balancing  -- we used all ,,syncope'' series, while the ,,nosyncope'' were selected at random to match the cardinality.
Then, we divided the balanced data into two sets:
\begin{itemize}
    \item training set (154 series) -- used to train the models,
    \item test set (38 series) -- used to evaluate the models.
\end{itemize}

\section{Quality measures}
\label{section:quality_measures}

\input{quality_measures/f_score_tfpn}

Additionally, since the classes were balanced, accuracy could be used as a quality measure.
\input{quality_measures/accuracy_anomalies}

\section{Experiments}
\label{section:experiments}

\begin{figure}
    \centering
    \includegraphics[width=\textwidth]{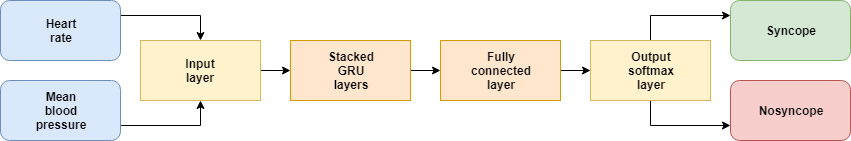}
    \caption{Neural network architecture}
    \label{fig:experiments:setup:architecture}
\end{figure}

As mentioned, we structured the syncope prediction problem as a classification task.
The overview of the used neural network architecture is presented in Fig.~\ref{fig:experiments:setup:architecture}.
Two main architecture variants involved using vanilla GRU layers and the bidirectional variant.
The loss used during training was categorical cross-entropy, with the batch size equal to \num{16} and ADADELTA optimizer.
The softmax output was compared with the threshold to select the final label.
We optimized the threshold value during the experiments.

\subsection{Hyperparameters optimization}

\begin{figure}
    \centering
    \includegraphics[width=\textwidth]{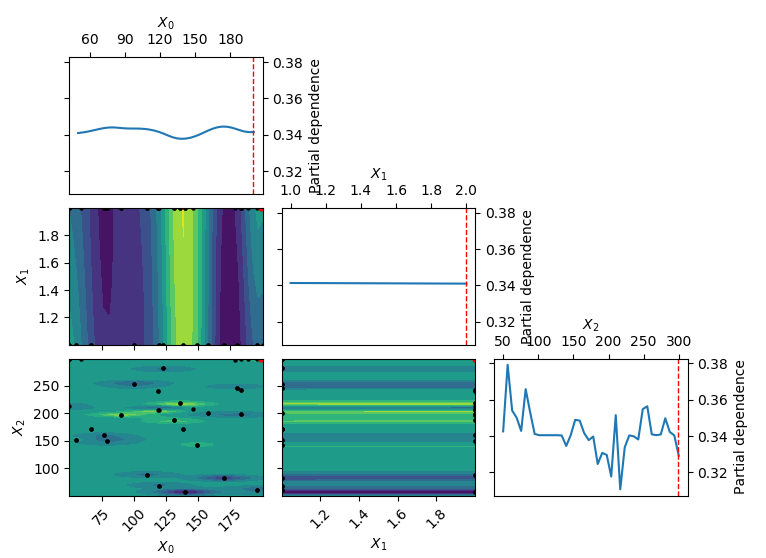}
    \caption{Partial dependence plots showing the relations between the number of GRU units in layer ($x_0$), the number of GRU layers ($x_1$) and history window size ($x_2$) and their influence on classification error (line charts).}
    \label{fig:experiments:params}
\end{figure}

The hyperparameters optimization, using Bayesian optimization, was conducted in two phases.
The first one aimed to determine which among the parameters have the most significant impact on proposed model quality.
There were seven parameters tested: the number of GRU units, the number of GRU layers, history window size, batch size, learning rate, learning rate decay, and output threshold.
The experiments showed that:
\begin{itemize}
    \item the relevance of the number of layers decreases when the number of unit increases,
    \item models tend to perform better when with smaller batch size,
    \item the output threshold should be no higher than 0.75.
\end{itemize}

The second phase focused on the most influential of the parameters: the number of GRU units, the number of GRU layers a history window size. Fig.~\ref{fig:experiments:params} shows the partial dependence plot for those parameters.

\subsection{Vanilla GRU}

\begin{table}
    \centering
    \begin{tabular}{lcccc}
        \toprule
         & \multicolumn{2}{c}{threshold $=0.7$} & \multicolumn{2}{c}{best optimized threshold} \\
         \cmidrule(r){2-3}\cmidrule(l){4-5}
         & F\textsubscript{1} score & accuracy & F\textsubscript{1} score & accuracy \\
        \midrule
        Single layer, 100 units & 0.667 & 0.526 & 0.750 & 0.789 \\ 
        Single layer, 200 units & 0.731 & 0.632 & 0.872 & 0.868 \\
        Two layers, 100 units each & 0.808 & 0.763 & 0.808 & 0.763 \\
        \bottomrule
    \end{tabular}
    \caption{F\textsubscript{1} score and accuracy obtained for three vanilla GRU architecture variants.}
    \label{tab:experiments:vanilla}
\end{table}

\begin{figure}
    \centering
    \includegraphics[width=\textwidth]{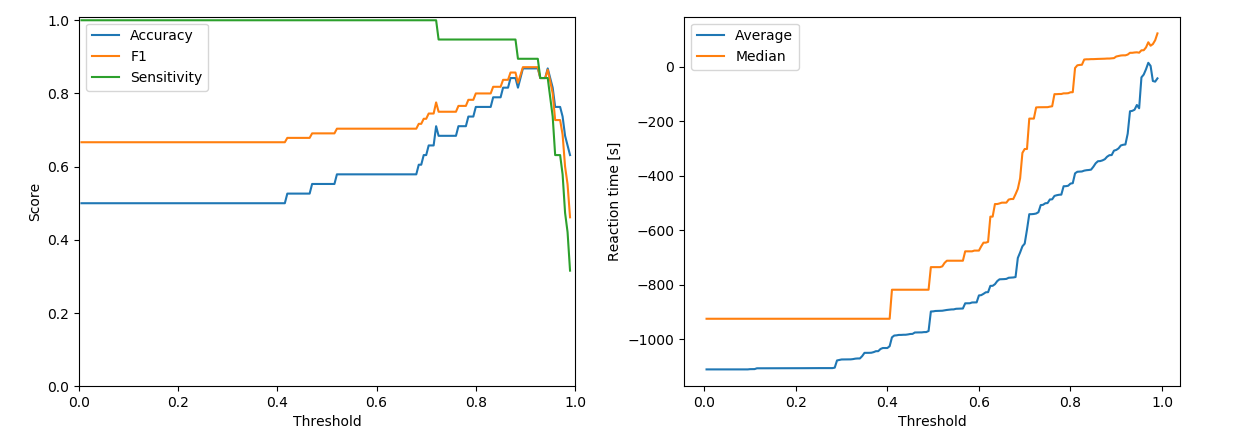}
    \caption{Relationships between threshold value and quality scores (on the left) and its influence on the syncope detection when compared to the manual marking (on the right) for the model with single vanilla GRU layer containing 200 units.}
    \label{fig:experiments:vanilla}
\end{figure}

All vanilla GRU models generated false negatives and did not significantly improve the reaction time when compared to the manually labeled time of the syncope.
Applying the per-model optimized threshold generally improved the accuracy and F\textsubscript{1} score results when compared to applying a fixed $threshold=0.7$ (Fig.~\ref{tab:experiments:vanilla}).
However, when one considers the recall (sensitivity) and the reaction time, the lower thresholds tended to yield better results (Fig.~\ref{fig:experiments:vanilla}).

\subsection{Bidirectional GRU}

\begin{table}
    \centering
    \begin{tabular}{lcccc}
        \toprule
         & \multicolumn{2}{c}{threshold $=0.7$} & \multicolumn{2}{c}{best optimized threshold} \\
         \cmidrule(r){2-3}\cmidrule(l){4-5}
         & F\textsubscript{1} score & accuracy & F\textsubscript{1} score & accuracy \\
        \midrule
        Single layer, 100 units & 0.745 & 0.658 & 0.800 & 0.763 \\ 
        Single layer, 200 units & 0.792 & 0.737 & 0.826 & 0.789 \\
        Two layers, 100 units each & 0.905 & 0.895 & 0.905 & 0.895 \\
        \bottomrule
    \end{tabular}
    \caption{F\textsubscript{1} score and accuracy obtained for three bidirectional GRU architecture variants.}
    \label{tab:experiments:bidirectional}
\end{table}

\begin{figure}
    \centering
    \includegraphics[width=\textwidth]{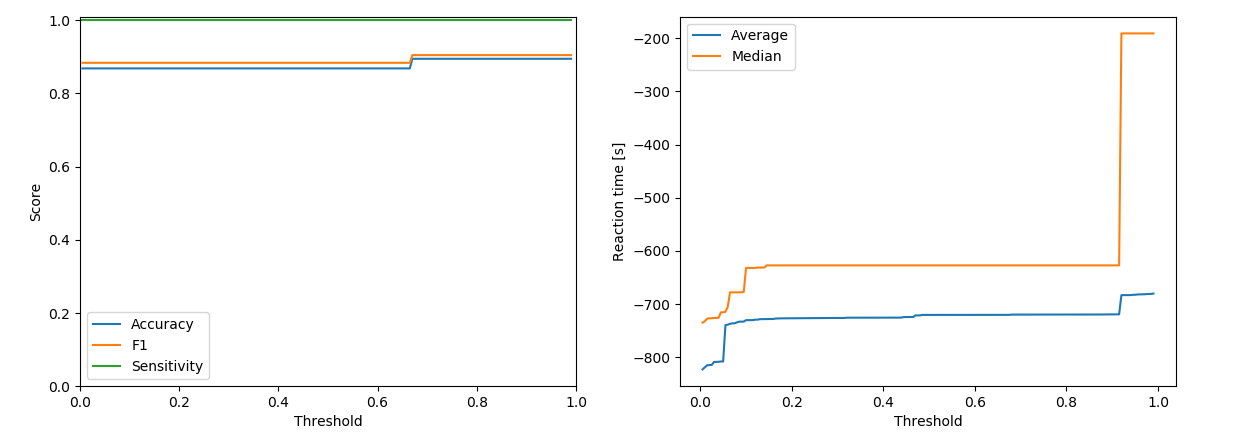}
    \caption{Relationships between threshold value and quality scores (on the left) and its influence on the syncope detection when compared to the manual marking (on the right) for the model with two bidirectional GRU layers containing 100 units each.}
    \label{fig:experiments:bidirectional}
\end{figure}

In the case of bidirectional GRU, applying the per-model optimized threshold also improved the results when compared to applying a fixed $threshold=0.7$ (see Tab.~\ref{tab:experiments:bidirectional}). However, the gains were not as significant as in the vanilla GRU case.
Moreover, as can be seen in Fig.~\ref{fig:experiments:bidirectional}, in case of the best of considered models, change of the threshold value has almost no impact on model quality.
Additionally, only the extreme values significantly impact the reaction time.

%% file: quality_measures/f_score_tfpn.tex
An F-measure was used as a quality metric during conducted experiments. It is calculated using two helper metrics, a recall (\ref{eq:recall}), also called sensitivity, and a precision (\ref{eq:precision}):

\begin{equation}
    \mathrm{recall} = \frac{\mathit{tp}}{\mathit{tp} + \mathit{fn}},
    \label{eq:recall}
\end{equation}
\begin{equation}
    \mathrm{precision} = \frac{\mathit{tp}}{\mathit{tp} + \mathit{fp}},
    \label{eq:precision}
\end{equation}
where:
\begin{itemize}
    \item $\mathit{tp}$ -- true positive -- item correctly classified as an anomaly,
    \item $\mathit{fp}$ -- false positive -- item incorrectly classified as an anomaly,
    \item $\mathit{fn}$ -- false negative -- item incorrectly classified as a part of normal operation.
\end{itemize} 

The $\beta$ parameter controls the recall importance in relevance to the precision when calculating an F-measure:

\begin{equation}
    \mathrm{F}_\beta = (1 + \beta^2) \cdot \frac{\mathrm{recall} \cdot \mathrm{precision}}{\mathrm{recall} + \beta^2 \cdot \mathrm{precision}}.
    \label{eq:f_measure}
\end{equation}

During the experiments $\beta = 1$ was used.

%% file: quality_measures/accuracy_anomalies.tex
Given the values $t$ and $f$ representing, respectively, an amount of correctly and incorrectly classified samples, the accuracy can be defined as in (\ref{eq:accuracy}):

\begin{equation}
\mathrm{accuracy} = \frac{\mathit{t}}{\mathit{t} + \mathit{f}}.
\label{eq:accuracy}
\end{equation}

%% file: contents-40-conclusions-future.tex
\section{Conclusions and future work}
\label{section:conclusions}

This work is preliminary research which addresses several issues related to falls prediction such as data preparation and model training. The best bidirectional GRU model enabled detection of forthcoming fall approx ten minutes before the event with approx 90 \% accuracy. It is worth noting the model is well suited for implementation in wearable divides with appropriate compression \cite{wielgosz2019mapping}. As future work, the authors are going to focus on enhancing the vanilla GRU solution with more advanced mechanisms and its architecture modifications for better performance. 

%% file: ack.tex
\section*{Acknowledgment}

This project was realized in collaboration with the Laboratory for Gravitational Physiology, Aging and Medicine Research Unit, Institute of Physiology, the Medical University of Graz, and with the cooperation of Department of Health Sciences and Information technologies at Alma Mater Europea University, Maribor, Slovenia. We would like to especially thank professor Nandu Goswami, who provided the data and explanations used to train and evaluate models.

%% file: article.bbl
\begin{thebibliography}{7}
\providecommand{\natexlab}[1]{#1}
\providecommand{\url}[1]{\texttt{#1}}
\expandafter\ifx\csname urlstyle\endcsname\relax
  \providecommand{\doi}[1]{doi: #1}\else
  \providecommand{\doi}{doi: \begingroup \urlstyle{rm}\Url}\fi

\bibitem[Radzio(2019)]{radzio2019falls}
Marcin Radzio.
\newblock Falls prediction with recurrent neural networks.
\newblock Master's thesis, AGH University of Science and Technology, 2019.

\bibitem[Martínez-Velilla et~al.(2015)Martínez-Velilla, Casas-Herrero,
  Zambom-Ferraresi, Suárez, Alonso-Renedo, Contín, de~Asteasu, Echeverria,
  Lázaro, and Izquierdo]{martinezvelilla2015functional}
Nicolás Martínez-Velilla, Alvaro Casas-Herrero, Fabrício Zambom-Ferraresi,
  Nacho Suárez, Javier Alonso-Renedo, Koldo~Cambra Contín, Mikel López-Sáez
  de~Asteasu, Nuria~Fernandez Echeverria, María~Gonzalo Lázaro, and Mikel
  Izquierdo.
\newblock Functional and cognitive impairment prevention through early physical
  activity for geriatric hospitalized patients: study protocol for a randomized
  controlled trial.
\newblock \emph{BMC Geriatrics}, 15\penalty0 (1), 2015.
\newblock ISSN 1471-2318.
\newblock \doi{10.1186/s12877-015-0109-x}.

\bibitem[Martínez-Velilla et~al.(2019)Martínez-Velilla, Casas-Herrero,
  Zambom-Ferraresi, de~Asteasu, Lucia, Galbete, García-Baztán, Alonso-Renedo,
  González-Glaría, Gonzalo-Lázaro, Apezteguía~Iráizoz,
  Gutiérrez-Valencia, Rodríguez-Mañas, and
  Izquierdo]{martinezvelilla2019effect}
Nicolás Martínez-Velilla, Alvaro Casas-Herrero, Fabricio Zambom-Ferraresi,
  Mikel López-Sáez de~Asteasu, Alejandro Lucia, Arkaitz Galbete, Agurne
  García-Baztán, Javier Alonso-Renedo, Belen González-Glaría, María
  Gonzalo-Lázaro, Itziar Apezteguía~Iráizoz, Marta Gutiérrez-Valencia,
  Leocadio Rodríguez-Mañas, and Mikel Izquierdo.
\newblock {Effect of Exercise Intervention on Functional Decline in Very
  Elderly Patients During Acute Hospitalization: A Randomized Clinical Trial}.
\newblock \emph{JAMA Internal Medicine}, 179\penalty0 (1):\penalty0 28--36, Jan
  2019.
\newblock ISSN 2168-6106.
\newblock \doi{10.1001/jamainternmed.2018.4869}.

\bibitem[Rajagopalan et~al.(2017)Rajagopalan, Litvan, and
  Jung]{rajagopalan2017fall}
Ramesh Rajagopalan, Irene Litvan, and Tzyy-Ping Jung.
\newblock Fall prediction and prevention systems: Recent trends, challenges,
  and future research directions.
\newblock \emph{Sensors}, 17\penalty0 (11), 2017.
\newblock ISSN 1424-8220.
\newblock \doi{10.3390/s17112509}.

\bibitem[Mateen et~al.(2016)Mateen, Bussas, Doogan, Waller, Saverino,
  Kir{\'{a}}ly, and Playford]{mateen2016machine}
Bilal~A. Mateen, Matthias Bussas, Catherine Doogan, Denise Waller, Alessia
  Saverino, Franz~J. Kir{\'{a}}ly, and E.~Diane Playford.
\newblock Machine learning in falls prediction; {A} cognition-based predictor
  of falls for the acute neurological in-patient population.
\newblock \emph{CoRR}, abs/1607.07751, 2016.

\bibitem[Nait~Aicha et~al.(2018)Nait~Aicha, Englebienne, Van~Schooten,
  Pijnappels, and Kröse]{naitaicha2018deep}
Ahmed Nait~Aicha, Gwenn Englebienne, Kimberley~S. Van~Schooten, Mirjam
  Pijnappels, and Ben Kröse.
\newblock Deep learning to predict falls in older adults based on daily-life
  trunk accelerometry.
\newblock \emph{Sensors}, 18\penalty0 (5), 2018.
\newblock ISSN 1424-8220.
\newblock \doi{10.3390/s18051654}.

\bibitem[Wielgosz and Karwatowski(2019)]{wielgosz2019mapping}
Maciej Wielgosz and Michał Karwatowski.
\newblock Mapping neural networks to {FPGA}-based {IoT} devices for ultra-low
  latency processing.
\newblock \emph{Sensors}, 19\penalty0 (13), 2019.
\newblock ISSN 1424-8220.
\newblock \doi{10.3390/s19132981}.

\end{thebibliography}
